\documentclass[letterpaper, 10 pt, conference]{ieeeconf}  

\IEEEoverridecommandlockouts                              

\usepackage{amssymb}
\usepackage{amsmath}
\usepackage{graphicx}
\usepackage{caption}
\usepackage{subcaption}
\graphicspath{{images/}}
\usepackage{float}
\usepackage{hyperref}
\newcommand{\norm}[1]{\left\lVert#1\right\rVert} 

\newcommand{\R}{\mathbb{R}}

\newcommand{\x}{\mathbf{x}}

\newcommand{\uu}{\mathbf{u}}
\newcommand{\ua}{\mathbf{a}}

\newcommand{\s}{\mathbf{s}}

\title{\LARGE \bf
Towards Standardized Disturbance Rejection Testing of Legged Robot Locomotion with Linear Impactor: A Preliminary Study, \\ Observations, and Implications}

\author{Bowen Weng$^{1}$, Guillermo A. Castillo$^{2}$,  Yun-Seok Kang$^{3}$, and Ayonga Hereid$^{4}$
\thanks{This work was supported in part by the National Science Foundation
under grant FRR-21441568 and in part by National Highway Traffic Safety Administration (NHTSA) U.S. Department of Transportation.}%
\thanks{A modified version of this preprint has been accepted at IEEE International Conference on Robotics and Automation (ICRA) 2024}%
\thanks{$^{1}$Department of Computer Science, Iowa State University, IA, USA;  {\tt\footnotesize bweng@iastate.edu.}}
\thanks{$^{2}$Electrical and Computer Engineering, The Ohio State University, Columbus, OH, USA;  {\tt\footnotesize castillomartinez.2@osu.edu.}}
\thanks{$^{3}$School of Health and Rehabilitation Sciences, The Ohio State University, Columbus, OH, USA. {\tt\footnotesize yunseok.kang@osumc.edu.}}
\thanks{$^{4}$Mechanical and Aerospace Engineering, The Ohio State University, Columbus, OH, USA. {\tt\footnotesize hereid.1@osu.edu.}}%
}

\begin{document}

\maketitle

\begin{abstract}
Dynamic locomotion in legged robots is close to industrial collaboration, but a lack of standardized testing obstructs commercialization. The issues are not merely political, theoretical, or algorithmic but also physical, indicating limited studies and comprehension regarding standard testing infrastructure and equipment. For decades, the approaches we have been testing legged robots were rarely standardizable with hand-pushing, foot-kicking, rope-dragging, stick-poking, and ball-swinging.
This paper aims to bridge the gap by proposing the use of the linear impactor, a well-established tool in other standardized testing disciplines, to serve as an \emph{adaptive}, \emph{repeatable}, and \emph{fair} disturbance rejection testing equipment for legged robots. A pneumatic linear impactor is also adopted for the case study involving the humanoid robot Digit. Three locomotion controllers are examined, including a commercial one, using a walking-in-place task against frontal impacts. The statistically best controller was able to withstand the impact momentum ($26.376$ $\mathrm{kg}\cdot\mathrm{m/s}$) on par with a reported average effective momentum from straight punches by Olympic boxers ($26.506 \mathrm{kg}\cdot\mathrm{m/s}$). Moreover, the case study highlights other anti-intuitive observations, demonstrations, and implications that, to the best of the authors' knowledge, are first-of-its-kind revealed in real-world testing of legged robots.
\end{abstract}

\section{Introduction}\label{sec:introduction}
Dynamic locomotion, i.e. the capability to create statistically stable and adaptive walking gaits that can withstand a certain degree of external disruption, has been a central pursuit for legged robot researchers for many decades~\cite{hobbelen2007disturbance,diedam2008online,posa2017balancing,meduri2023biconmp,chen2023quadruped,castillo2021robust}. The complex, articulated structure of mechanical limbs, coupled with the prevalent under-actuation and the uncertainties encountered in real-world environments, present a distinct and critical challenge in the field of legged robotics. This challenge distinguishes legged robots from other mobility alternatives, such as wheeled robots, in its uniqueness and complexity.

Fortunately, the past decades have also observed a tremendous amount of proposals tackling the locomotion challenge against external disturbances utilizing complex approaches involving model insights~\cite{diedam2008online,posa2017balancing,meduri2023biconmp,chen2023quadruped}, machine learning~\cite{miki2022learning,hwangbo2019learning,castillo2021robust,radosavovic2023learning}, and various combinations of the two. Many of the proposals also demonstrate real-world operations or in the wild. However, there still is a large gap towards scaling up the deployment of legged robots to work alongside humans in an array of industrial and commercial environments. One of the main bottlenecks lies with the lack of formal and unbiased performance testing standards and certificates. To a very large extent, even under specific functionalities such as the disturbance rejection domain with fallover being the only failure event of concern, one does not necessarily know if the robot ``works", ``works well", or ``works well enough". The remediate of this gap requires a series of cross-disciplinary work addressing the theoretical foundation of testing~\cite{weng2023comparability}, the testing algorithms~\cite{weng2022safety}, and the development of testing equipment and infrastructure~\cite{aller2019state}. In particular, the last topic has been significantly overlooked and understudied in the legged robot field and is also the primary focus of this paper. 

For the past few decades, \emph{researchers have been largely relying on non-rigorous, non-repeatable, and occasionally unfair testing apparatus for the conceptual performance demonstration of legged robots against external disturbances}. We used hand-pushing~\cite{castillo2021robust}, foot-kicking, rope-dragging~\cite{radosavovic2023learning}, hockey sticks poking, yoga-ball-throwing~\cite{radosavovic2023learning}, and swing-ball impact~\cite{chen2023quadruped}. Despite their widespread use, intuitive (and somewhat impressive) demonstrations, these approaches are barely a solid test of the robots' true capabilities as they lack standardization, consistency, and rigorous flexibility. The researchers have also performed simulation-oriented tests generating the external impact as forces posed upon a certain contact point on the robot~\cite{castillo2021robust,weng2022safety}. Those forces are often applied to the robot in an unrealistic manner. For example, the direction of the external force often remains still regardless of the orientation and position variations of the robot body during the contacting period after impact. The forces are also configured to exhibit some ``ideal" patterns, such as Quasi-static, periodic, and spiking patterns. All of the above require accurate control of the impact time and impact duration, which would not be possible in real-world practice against intelligent legged robots, and could also lead to unfair tests as one will clarify later in Section~\ref{sec:case}.

\subsection{Main Contributions}
This paper introduces a \emph{novel proposal to utilize an established piece of impact generation equipment}, specifically referred to as the \emph{linear impactor}~\cite{walsh2011influence,allison2015measurement,kang2023comparison,edwards2023kinematic}, for the standardized testing of legged robot locomotion against external physical disturbances such as impacts\footnote{Related video: \href{https://youtu.be/bUWnnTYNL3A}{https://youtu.be/bUWnnTYNL3A}}. This adaptation positions the linear impactor as a candidate for creating rigorous, objective, and repeatable performance tests, enabling the assessment of legged robots in a manner that is both fair and adaptable. To our knowledge, this study represents the first effort within the field of legged robotics to address the need for suitable testing equipment and infrastructure, focusing on the aspects that ensure unbiased and consistent evaluation.

Moreover, a \emph{real-world in-lab experiment} was conducted to assess the disturbance rejection performance of the Digit humanoid robot. This involved using a pneumatic linear impactor to subject the robot, equipped with three different state-of-the-art locomotion controllers, to various physical disturbances. Notably, this same pneumatic impactor was previously employed in studies on injury biomechanics~\cite{kang2023comparison}, head and helmet impacts~\cite{allison2015measurement}, underscoring its utility across diverse applications. The experiment unveiled several unique insights and challenges specific to the testing of legged robots. Among these findings was the anti-intuitive observation that a more severe impact does not necessarily translate into a worse outcome. This is attributable to the complex and intelligent feedback mechanisms inherent in legged robot locomotion control, a phenomenon that had only been identified in simulation-based studies until now~\cite{weng2022safety, weng2023comparability}.

\section{Preliminaries and Problem Formulation}\label{sec:prob}

\subsection{Problem Formulation}
The testing subject in this study is a certain legged robot capable of achieving the functionality of dynamic locomotion. It essentially admits the system formulations as
\begin{equation}\label{eq:subject_ctrl_sys}
    \s(t+1) = h(\s(t),\uu(t); \omega(t)).
\end{equation}
with state $\s \in S \subset \R^n$, action $\uu \in \mathcal{U} \subset \R^m$, and uncertainties $\omega \in W \subset \R^{w}$. The discrete-time characterization suits the practical needs with testing execution and data acquisition primarily involving digital equipment. 

Note the uncertainties are not involving any environmental factors, but are primarily internal such as the varying frictions of joints, the latency in the controller, and sensor noise. Note this approach differs from a typical stochastic system formulation for legged robots. In the context of testing, the so-called environmental disturbances are not randomly presented but are explicitly given and controlled with precision. This deliberate manipulation contrasts with traditional methods, allowing for more accurate assessments and understandings of the system's response to specific disturbances. As a result, the control action $\uu$ includes two major components as 
\begin{equation}
    \uu = \begin{bmatrix}
        \uu_i & \uu_e
    \end{bmatrix}.
\end{equation}
The $\uu_i$ denotes the typical control signal that is part of the testing subject such as the commended torques at all actuators (hence the subscript ``i" highlights the ``internal" nature). On the other hand, $\uu_e$ denotes the environmental factors and external disturbances such as the wind speed, the push-over action, the presence of other obstacles, and the room temperature. Those are the essential factors to be controlled in the context of performance testing.

As the focus of this paper is the disturbance rejection performance against external impact, the primary component of $\uu_e$ is thus a certain notion of the physical ``impact". Other factors will also be rigorously controlled and monitored to ensure they do not pose significant effects on the accuracy and repeatability of the testing outcomes.

In the practice of tests, the impact action $\uu_e$ is typically generated through a certain testing equipment operated by human engineers including the physical apparatus and other control and monitoring software modules. Formally speaking, we have the determination of $\uu_e$ satisfying
\begin{equation}\label{eq:testing_ctrl}
    \uu_e(t) = g(\x(t), \ua(t);\omega_g(t)),
\end{equation}
with state $\x \in X \subset \R^x$, action $\ua \in \mathcal{A} \subset \R^a$, and uncertainties $\omega_g \in W_g \subset \R^{w_g}$. The $\ua$ denotes the direct action one can take as a human testing operator (e.g., the releasing height of the pendulum impactor and the air pressure of the pneumatic impactor) for controlling the testing equipment. The state space $X \subset S$ as it is primarily concerned with the subject features directly related to the testing action propagation (e.g., the specific status of joints and actuators are not of primary concern for the impact testing equipment hence they are not part of $\x$).

Moreover, the internal control actions are determined by a certain locomotion control policy
\begin{equation}\label{eq:subject_ctrl}
    \uu_i(t) = \pi(\s(t)).
\end{equation}
Composing \eqref{eq:subject_ctrl} with \eqref{eq:subject_ctrl_sys}, one has
\begin{equation}\label{eq:subject_sys}
    \begin{aligned}
        \s(t+1) & = h(\s(t),\begin{bmatrix} \pi(\s(t)) & \uu_e(t) \end{bmatrix}; \omega(t)) \\
        & = f(\s(t), \uu_e(t); \omega(t)).
    \end{aligned}    
\end{equation}
Note the testing also involves the black-box system nature~\cite{weng2023comparability} with $f \in \mathcal{F}$ and $\mathcal{F}=S^{S \times \mathcal{U}_e \times W}$ being the set of all possible mappings. This has also been concretely observed in practice given the variety of legged robot hardware mechanism designs and software controller development. Note the unknownability and the stochasticity are both part of the subject nature and should be respected by the testing perspective.

The purpose of this paper is to present a certain testing equipment with system $g$ as defined by~\eqref{eq:testing_ctrl} such that it is
\begin{enumerate}
    \item \textbf{Adaptive}: The generated $\uu_e$ works with a variety of systems and robots $f \in \mathcal{F}$; 
    \item \textbf{Repeatable}: The system $g$ satisfies $W_g=\emptyset$ and for all $\x\in X$ and any pair of sufficiently close testing actions, $\uu_e^1$ and $\uu_e^2$ with $\norm{\uu_e^1-\uu_e^2}_p$ sufficiently small, $\norm{g(\x,\uu_e^1)-g(\x,\uu_e^2}_p$ is also sufficiently small;
    \item \textbf{Fair}: $\uu_e$ is strictly determined by $g$ for fixed state-actions, and should not be impacted by $h$ or $f$.
\end{enumerate}
Unfortunately, existing impact generating solutions with hands, feet, hockey sticks, and yoga balls for legged robot testing rarely satisfy the above properties. This paper is thus inspired to explore impact generation in a broader scope with lessons learned from other disciplines.

\subsection{Creating the Impact}
The notion of creating a physical ``impact" and identifying outcomes thereafter has been a commonly observed testing configuration throughout multiple disciplines with materials~\cite{meyers2008mechanical}, biomechanics~\cite{kang2023comparison,allison2015measurement}, consumer electronics~\cite{varghese2003test}, helmets~\cite{viano2012change}, to name a few.

One of the earliest impact tests is the Charpy Impact Test, also known as the Charpy V-notch test, Notched Bar Impact Testing, and Pendulum Impact Test, was first proposed in 1896~\cite{meyers2008mechanical}, to determine the amount of energy absorbed by a material during fracture. It has further extended to a variety of tests against helmets, footwear, electronics, etc. The impact was primarily generated with a pendulum of known mass and length that is dropped from a predetermined height. A similar setup called the Ball Impact Test (i.e. replacing the pendulum with a ball attached to a swigging string) has also been widely adopted in the glass industry to determine the strength and integrity of glass.

A significant limitation of the experimental setup described above, particularly when applied to the testing of legged robots, lies in its lack of flexible adaptability. Traditional testing methods, such as those employing pendulum impacts or swing-ball mechanisms, are often constrained by their reliance on gravity and the specific physical setup. These factors limit the impact momentum and speed that can be achieved, and are mostly comparable with small-scale and stationary subjects. Such gravity-dependent testing mechanisms present substantial challenges when attempting to adapt them to more complex scenarios, such as interactions with large-scale or moving subjects. The inherent design limitations restrict the ability to achieve impact motions with higher degrees of freedom, thereby reducing the test's relevance to real-world scenarios where multi-dimensional movements and forces are encountered. The same problem has also been observed with other impact generators such as the Shock Testing and Drop Tower Tests~\cite{burgin2007drop,cheng2010drop}.

\section{Testing Legged Robot Locomotion with the Linear Impactor}\label{sec:main}
\begin{figure}
    \vspace{3mm}
    \centering
    \includegraphics[width=0.99\linewidth]{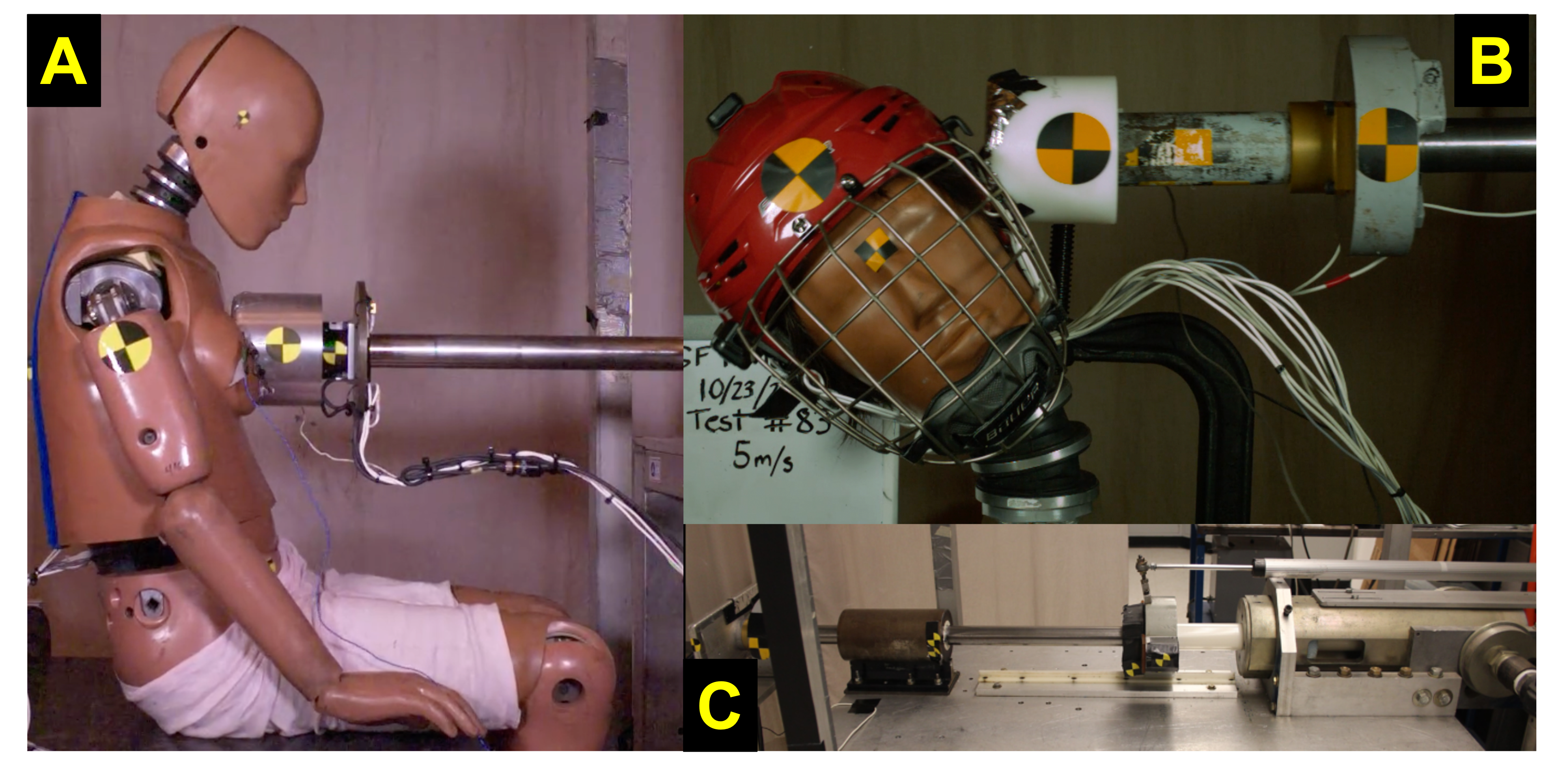}
    \caption{The linear pneumatic impactor has been used as a standard testing equipment related to the research in automobile safety and injury biomechanics simulating impact energy to head, tibia, thorax, abdomen, and shoulder impacts: A shows the test of dummy thorax impact, B is coming from the study of head and hockey helmet instrumentation evaluation~\cite{allison2015measurement}, both tests were performed with the linear pneumatic impactor with different impact energies shown in C. The same device is also configured with minor modifications for legged robot locomotion testing performed in this study.}
    \label{fig:impactor_example}
\end{figure}

In this section, the linear impactor is introduced as a proposed testing equipment towards a standardized disturbance rejection testing of legged robots.

\begin{figure}
    \centering
    \includegraphics[trim={0 9.5cm 0 0},clip,width=0.99\linewidth]{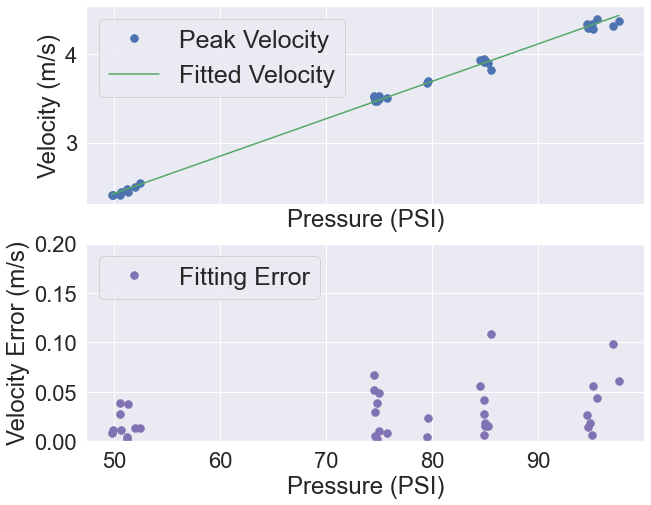}
    \caption{The pneumatic impactor adopted in this study demonstrates its repeatability by empirically showing a linear relation between the testing action $\uu_e$, the peak velocity achieved by the impactor, and the operator control $\ua$, the configured pressure for the compressed air. The linear fitting error is within 0.1 m/s.}
    \label{fig:repeatibility}
\end{figure}

\subsection{The Linear Impactor}
Intuitively, a linear impactor produces the impact motion by driving a certain object (commonly referred to as the impactor) along a linear path to impact the subject target. One of the common ways to initiate the impact is through compressed air with the corresponding impactor referred to as the pneumatic impactor. It has gained extensive applications in a variety of disciplines. The pneumatic impactor used in this study (see Fig.~\ref{fig:impactor_example}) has also been traditionally adopted as the standard testing equipment to support research activities related to vehicle occupant safety, injury biomechanics, and sports biomechanics~\cite{kang2023comparison,allison2015measurement}. In practice, there are other designs such as the Hydraulic Impactor~\cite{li2018development}, Mechanical Impactor, Motor-Driven Impactor, and Electromagnetic Impactor~\cite{mejia2019linear}, some of which are not necessarily confined to the testing discipline~\cite{li2018development}. 

The setup of the linear impactor allows the testing subject to be configured in an open space with sufficient room for extended motion. The impactor face is free to deflect and rebound in a natural way to assist the study of the complete impact kinematics. A variety of ram shafts, components, and impactor faces can be configured to achieve desired impact energies and simulated impact surfaces. The impactor can also be installed on other mobile platforms to achieve a higher degrees-of-freedom. Its \emph{adaptivity} is hence immediate given the flexible configurations mentioned above.

\subsection{Testing Configuration and Repeatability}
The primary class of subject robot being tested in this paper is the class of two-legged robots including the bipedal and the humanoids. Yet the methodologies can also be extended to other types of legged robots such as the quadruped. 

The robot could be configured with different designs revealed by the configuration of \eqref{eq:subject_ctrl_sys} and equipped with different locomotion algorithms formulated as \eqref{eq:subject_ctrl}, leading to a variety of testing subject. They are then tested with the testing control being the impact as described by~\eqref{eq:subject_sys}. Every ``test" thus involves a series of manipulations centered around the presented dynamics of~\eqref{eq:subject_sys}.

Generally speaking, every test starts by configuring the robot at a certain state $\s(0) = \s_0 \in S$. For the walking-in-place example, $\s_0$ characterizes the position, the desired walking gaits with zero velocity, and the orientation of the robot. As highlighted with \eqref{eq:subject_ctrl_sys}, the specific stance phase, swing phase, swing foot clearance at the moment of impact are essentially part of the uncertainties $\omega$. The diversities of those uncertainties should be respected by the testing program and it would be unfair to force the robot to be tested at a specific walking phase.

Reflected in the practice of testing, the robot needs to be brought or controlled into the desired $\s_0$ or a sufficiently small neighbourhood near $\s_0$. This is often with the assistant of a human testing operator. Note the human or external assistance is a common setup in the practice of testing. For example, the testing of Automated Driving Assist Systems also requires the expert testing driver or a so-called steering robot to bring the vehicle to a certain desired speed before starting the testing procedure~\cite{rao2019test}. The impact event is further initiated by configuring the operator action $\ua$ (e.g. the compressed air pressure for the pneumatic impactor used in the next section). The impact action $\uu_e$ is then propagated through the impactor equipment. 

Note the impact action in this study is the maximum velocity the impactor manages to achieve before the impact referred to as the \emph{peak velocity}. All tests are configured in the same way such that the peak velocity is achieved before the impact event. For the testing subject that is stationary, one can align the impact velocity (the velocity of the impactor at the moment of impact) with the peak velocity sufficiently well by placing the subject appropriately. This is not possible for the legged robot testing. The torso of the robot cannot remain absolutely still and the the robot cannot stay ``absolute" in-place even the desired velocity is set to zero. The variances exhibited in the walking patterns could be of significant diversities among the testing subject. They are components of the internal uncertainties characterized by $\omega$ in \eqref{eq:subject_ctrl_sys} and could also be the choice of the internal control signal $\uu_i$ determined by each individual locomotion controller~\eqref{eq:subject_ctrl}. The position drifts and torso shaking as mentioned above are empirically minor throughout all tests performed in this study (see Fig.~\ref{fig:AR85_02_logs} and Fig.~\ref{fig:AR95_01_logs} for example), yet their impact to the locomotion performance outcome is unclear.

This section is concluded with the repeatability demonstration (see Fig.~\ref{fig:repeatibility}) of the linear pneumatic impactor mentioned above and used throughout all tests discussed in the next section. The repeatability is demonstrated within the velocity range between 1.5 m/s and 4.4 m/s with the error of less than 0.1 m/s. The particular impactor is capable of reaching approximately 10 m/s velocity with the similar repeatability performance, yet that are beyond the current capability of the subject robot studied in this paper.

\section{A Case Study with Pneumatic Impactor and Walking-in-Place Testing} \label{sec:case}

\begin{figure*}
    \centering
    \vspace{3 mm}
    \includegraphics[width=0.99\linewidth]{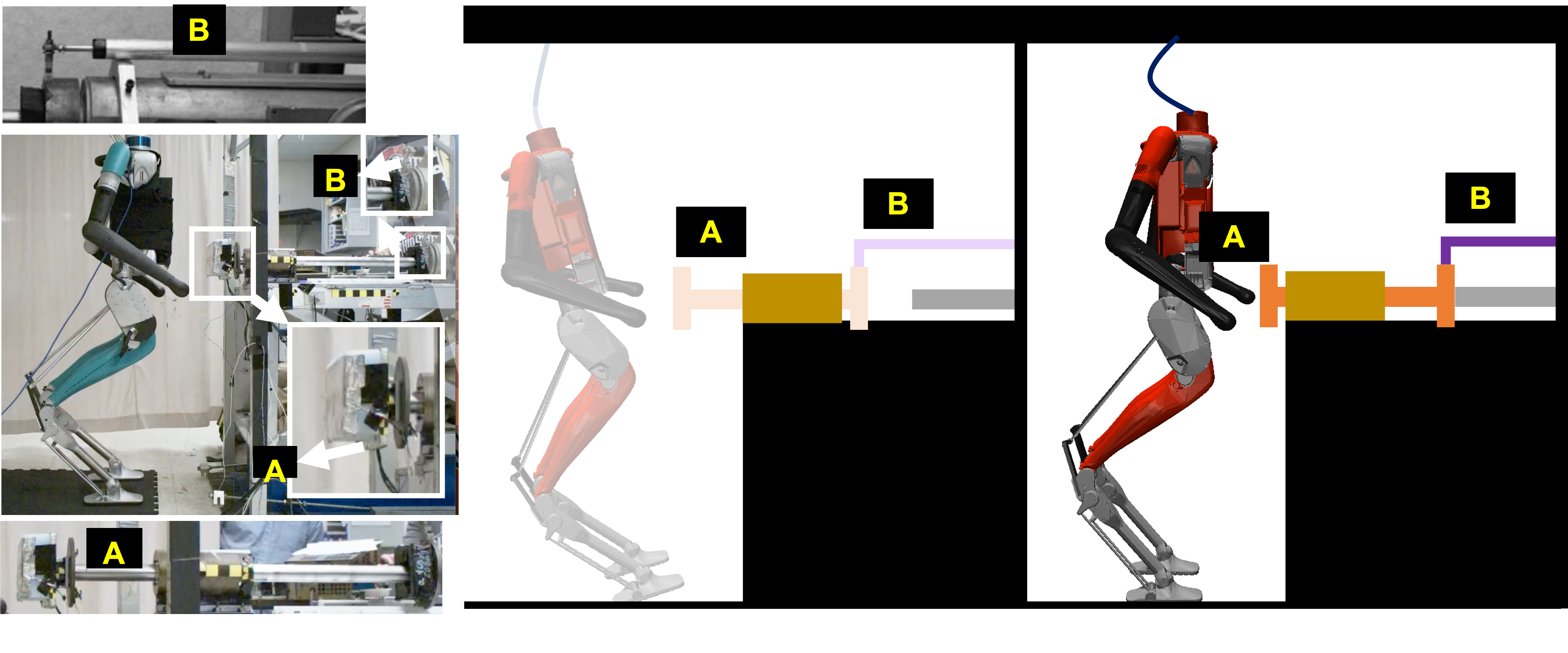}
    \caption{The testing configuration in-lab and the conceptual illustration before and after the impact: ``A" denotes the impactor face with a rectangular surface, ``B" denotes the displacement potentio-meter, one can also refer to Fig.~\ref{fig:impactor_example}C for another view of the same equipment.}
    \label{fig:configuration}
\end{figure*}
\begin{figure*}
    \centering
    \includegraphics[width=0.99\linewidth]{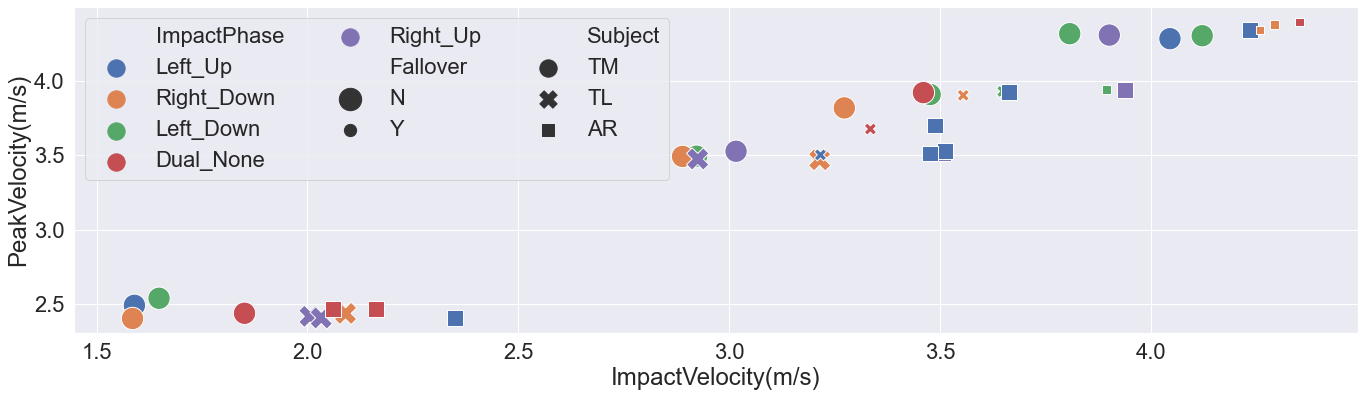}
    \caption{An overview of all 36 impact tests against three locomotion controllers: different impact phases are differentiated by the marker color and the label is specified in the form of stance foot phase (whether the left or the right foot is on the ground) and swing foot phase (whether the swing foot is going up or stepping down) with an underscore ``\_" in between. Note the dual support phase (with both feet on-ground) is specified as ``Dual\_None". The fallover status is characterized by the size of the marker. The three different locomotion controllers are differentiated by the marker type.}. 
    \label{fig:all_tests}
\end{figure*}

\begin{figure}
    \centering
    \includegraphics[width=0.45\textwidth]{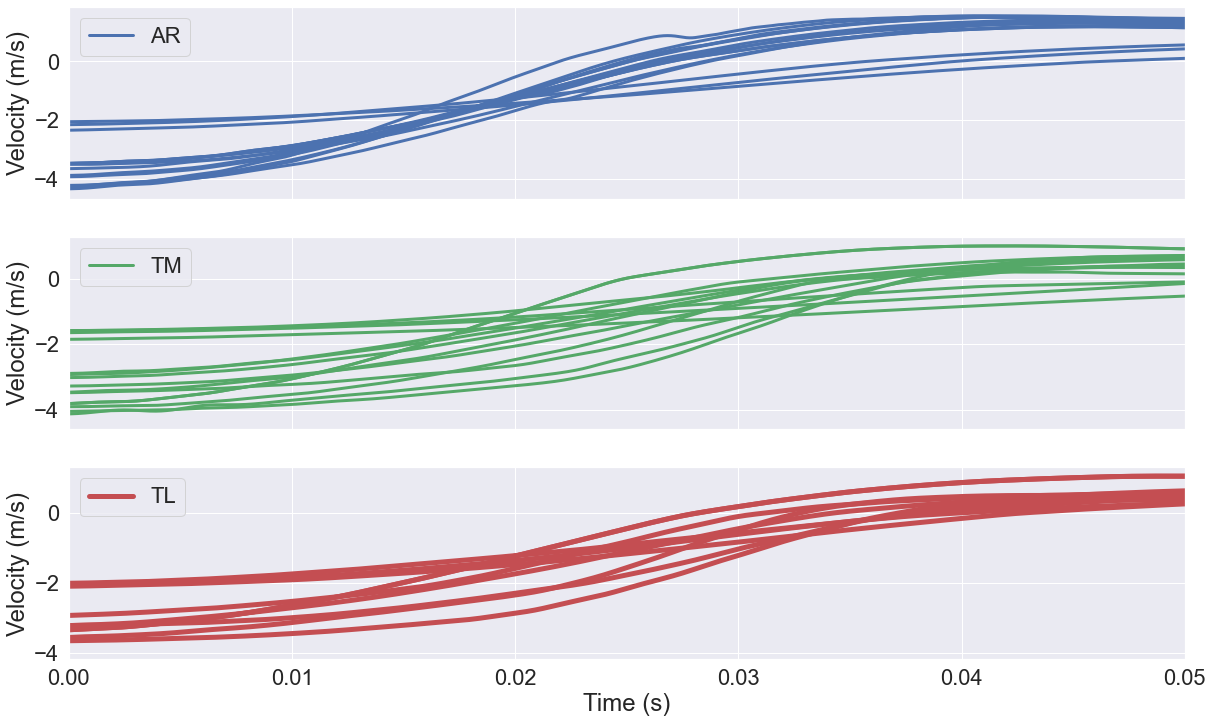}
    \caption{The velocity profiles of the impactor within 0.05-second after the moment of impact.}
    \label{fig:head_velocities}
\end{figure}

For the particular pneumatic impactor adopted in the case study section discussed next, the overall configuration is shown in Fig.~\ref{fig:configuration}. The center of the impactor face is 99.2 cm from the ground. The impactor weights 6.4 kg. The impactor face is a rectangle of 6-inch $\times$ 4-inch (about 15.24 cm $\times$ 10.16 cm). More details regarding the subject robot and testing configurations are explained as follows.

\subsection{Subject Robot Configuration}
The same Digit robot is adopted throughout all impact tests. Three different locomotion controllers are tested, including (i) a template-model (TM) inspired whole body controller, (ii) a template-model based task space learning policy (TL)~\cite{castillo2023template}, and (iii) a previous version of the default controlled coming with Digit designed by its original manufacturer Agility Robotics (AR). Specifically, the TM controller is primarily based on the Task Space Inverse Dynamics (TSID) methodology introduced in~\cite{del2016implementing} with a high-level step planner based on Raibert's regulations~\cite{raibert1983dynamically}. The TL controller follows the hierarchical framework in~\cite{castillo2023template} that combines a reinforcement learning-based high-level planner policy for the online generation of task space commands with a low-level TSID controller to track the desired task space trajectories. The AR controller is part of a commercially available product, hence the technical details are not open and unclear. Those controllers are also different at the selected gait (e.g., stepping frequency and swing foot clearance), the desired torso height, and many other known, unknown, observable, and non-observable configurations. 

All of the controllers have demonstrated an overall good performance for stepping-in-place, disturbance rejection, walking over various terrain conditions, and in the wild. However, it remains unclear to what concrete extent the locomotion control is stable and safe (i.e. performance characterization), which one is the best (i.e. benchmark comparison), to name a few. 

\begin{figure*}
\begin{minipage}{.5\textwidth}
  \begin{subfigure}{\linewidth}
    \centering
    \includegraphics[width=.95\linewidth]{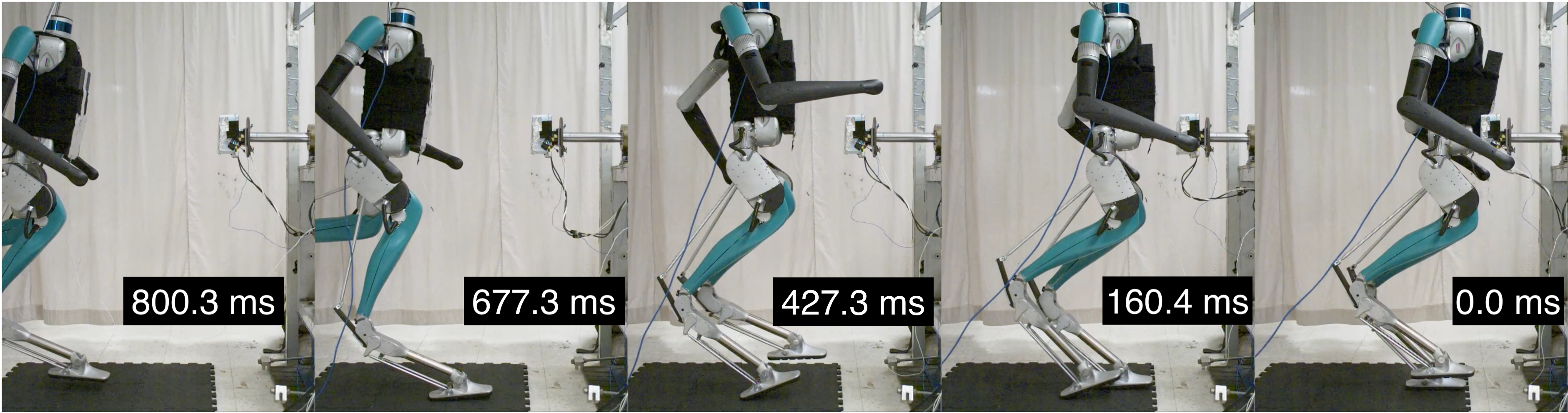}
    \caption{}
    \label{fig:AR85_02_frames}
  \end{subfigure}\\[1ex]
  \begin{subfigure}{\linewidth}
    \centering
    \includegraphics[width=.95\linewidth]{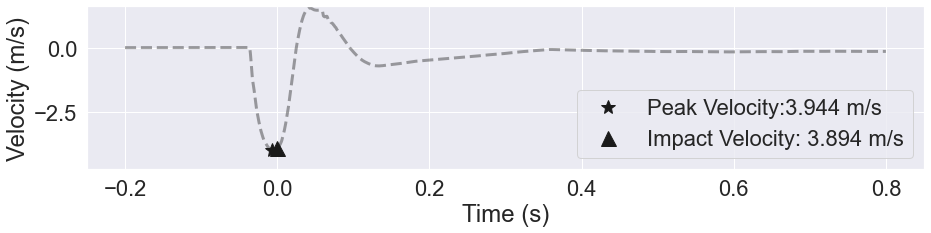}
    \caption{}
    \label{fig:AR85_02_headvel}
  \end{subfigure}
\end{minipage}%
\begin{minipage}{.5\textwidth}
    \begin{subfigure}{\linewidth}
        \centering
        \includegraphics[width=.95\linewidth]{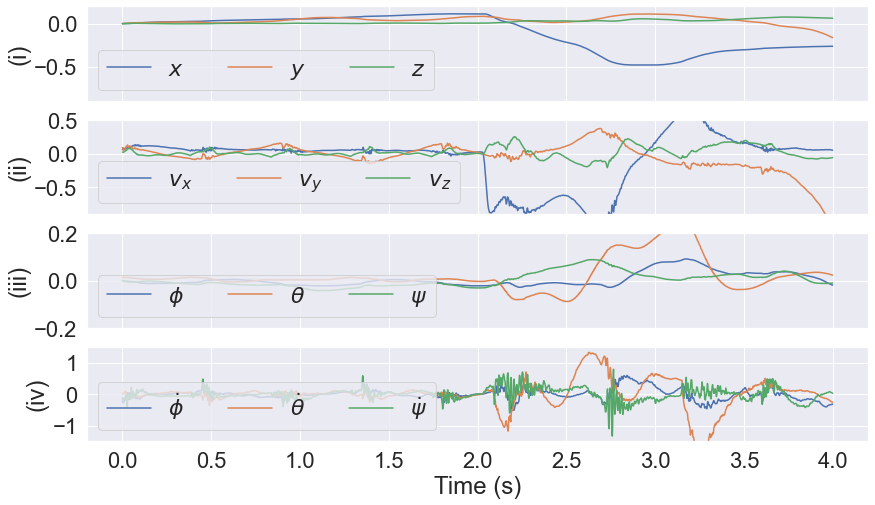}
        \caption{}
        \label{fig:AR85_02_logs}
    \end{subfigure}
\end{minipage}
\caption{The AR-equipped Digit robot falls over against the frontal impact velocity at 3.89 m/s: \ref{fig:AR85_02_frames} denotes the frames captured by the high-speed camera at and after the moment of impact (from right to left), \ref{fig:AR85_02_headvel} illustrates the velocity profile of the impactor within the time range in between 0.2 s before the impact and 0.8 s after the impact, \ref{fig:AR85_02_logs} shows the robot dynamic states in a 4-second window including the impact moment, the first row denotes center of mass (CoM) position of the torso, the second row denotes the CoM velocity, the third row shows the roll ($\phi$), pitch ($\theta$), and yaw ($\psi$) angles of the robot torso, and the last row shows the roll rate, pitch rate, and yaw rate of the robot torso.}
\label{fig:AR85_02}
\vspace{-2mm}
\end{figure*}

\begin{figure*}
\begin{minipage}{.5\textwidth}
  \begin{subfigure}{\linewidth}
    \centering
    \includegraphics[width=.95\linewidth]{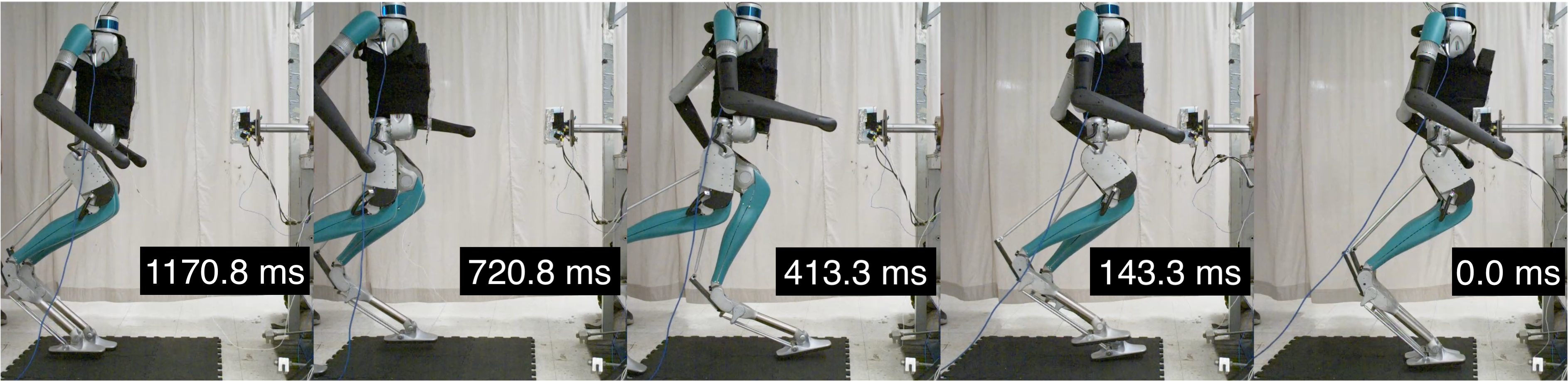}
    \caption{}
    \label{fig:AR95_01_frames}
  \end{subfigure}\\[1ex]
  \begin{subfigure}{\linewidth}
    \centering
    \includegraphics[width=.95\linewidth]{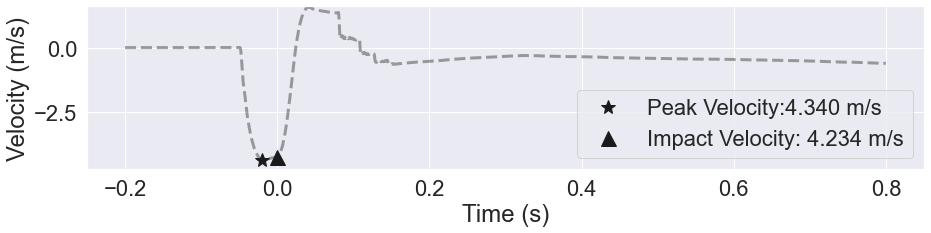}
    \caption{}
    \label{fig:AR95_01_headvel}
  \end{subfigure}
\end{minipage}%
\begin{minipage}{.5\textwidth}
    \begin{subfigure}{\linewidth}
        \centering
        \includegraphics[width=.95\linewidth]{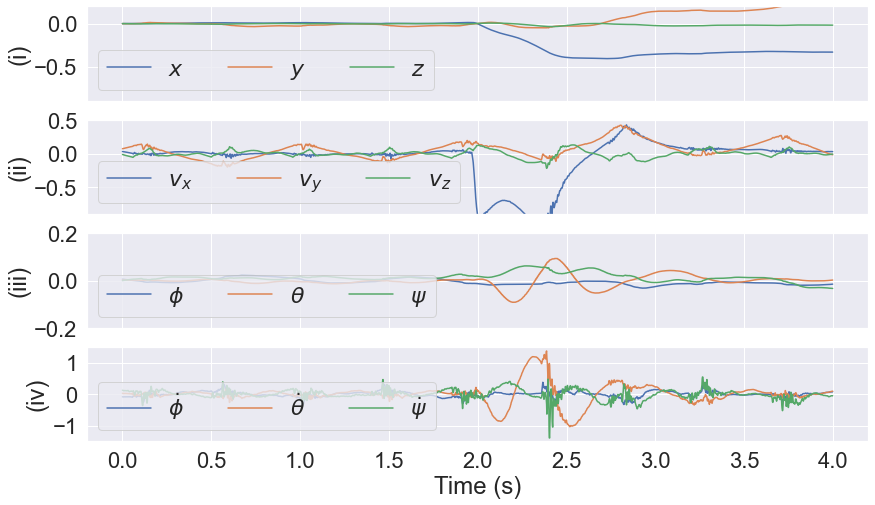}
        \caption{}
        \label{fig:AR95_01_logs}
    \end{subfigure}
\end{minipage}
\caption{The AR-equipped Digit robot recovers from the frontal impact velocity at 4.23 m/s (a more significant impact compared with Fig.~\ref{fig:AR85_02}): the sub-figures \ref{fig:AR95_01_frames},\ref{fig:AR95_01_headvel}, and \ref{fig:AR95_01_logs} share a similar setup as described in the caption of Fig.~\ref{fig:AR85_02}.}
\label{fig:AR95_01}
\end{figure*}

\subsection{Testing Configuration}
For start, as discussed in Section~\ref{sec:main}, the robot is controlled to achieve the desired walking-in-place motion at a fixed point near the impactor where the head is expected to reach its peak velocity (whose repeatability has been demonstrated in Fig.~\ref{fig:impactor_example}). Such a controlled initialization is achieved in a variety of ways based on the functional nature of each locomotion controller. 

For TM and TL, the robot is brought to the fixed point with the assistance of the human testing operator and the walking-in-place is triggered by configuring the desired velocity to zero (yet one may still drift as mentioned above in Section~\ref{sec:main}). For AR, the controller is technically not capable of walking-in-place as the robot will simply stop and stand still if the the desired velocity is set to zero. As a result, the walking-in-place motion for AR is achieved with the human operator's help through consecutively giving small desired speed signals ($\leq 0.1$ m/s) to keep the robot close to the fixed point. For all impact tests performed, the robot remains within a short distance near the fixed point. This can be demonstrated by the short time gaps between the peak velocity and the identified impact velocity as shown in Fig.~\ref{fig:AR85_02_headvel} and Fig.~\ref{fig:AR95_01_headvel}. Specifically, among all 36 tests, the average time gap between reaching the peak velocity and the moment of impact is 0.0453 ($\pm$0.0342)~s. The AR oriented tests are with the shortest average time gap of 0.0184 ($\pm$0.0201)~s whereas the TM oriented tests tend to take longer for the impact to occur with the average gap time of 0.0722 ($\pm$0.0334)~s. This also highlights the inherent capability differences among different controllers in walking-in-place.

To help protect the robot, especially the sensors (one of the camera modules is near the estimated impact point), a Polyethylene foam block of 5.4 cm thickness are installed at the expected impact point and an Ethafoam block is installed on the impactor face. This can be viewed as part of the system \eqref{eq:subject_ctrl_sys} and remains configured in the same way for all tests against all subject controllers. As shown in Fig.~\ref{fig:configuration}, the robot is also loosely connected to a sliding rail installed above the robot. This is an added protection to the robot to prevent mechanical breaks and other damages caused by an actual fallover. As a result, the robot will not physically fallover (i.e. having any part of the robot physically touching the ground other than the two feet). The justification of whether a robot fails in the test is justified by combining the testing operator's observation and the post analyses of the dynamic states of robots after each test.

For each subject locomotion controller, the tests start with the controlled pressure around 50 PSI (Pounds per square inch) and one gradually increases the pressure to up to 95 PSI. If the subject controller fails (i.e., falls over) consecutively two trials, the tests will stop and one moves on to the next subject to prevent potential damages to the robot from repetitive failures. As a result, TL ends up with the smallest number of tests (9), TM and AR are tested with 13 and 14 impact events, respectively. Note real-world testing is a complex process. The initialization, calibration, reset, configuration, and many other standard procedures highlight the challenges of performing rigorous real-world tests which may have been overlooked by some existing works in the field. The 36 impact events are conducted for the purpose of a preliminary study with the proposal of this paper. Despite the limited number of tests, we can already draw some interesting observations and insights, which will be discussed next.


\subsection{Observations and Analyses}
An overview of the 36 impact events are summarized in Fig.~\ref{fig:all_tests}. The stepping phase at the moment of impact, the fallover outcome, and the subject controller involved in the test are also illustrated. 

There are two specific observations/insights one shall highlight in the section: (i) the necessity of characterizing the impact testing action $\uu_e$ as impact momentum (instead of force) to ensure the testing fairness, and (ii) the anti-intuitive observation where a more significant impact does not necessarily lead to a more severe outcome.

First, in the legged robot regime, the external force has been adopted extensively to characterize the impact disturbance, especially in the computer simulations~\cite{lee2022online,romualdi2022online,castillo2021robust,castillo2023template,weng2022safety}. This may not be a fair setup as the impact force is jointly determined by the impact momentum and impact duration. The impact duration is a changing variable that is not only determined by the testing equipment/algorithm, but more importantly, by the subject itself. As shown in Fig.~\ref{fig:head_velocities} (also partially revealed in Fig.~\ref{fig:AR85_02_headvel} and Fig.~\ref{fig:AR95_01_headvel}), throughout all tests, the impactor's velocity exhibits a similar pattern after the moment of impact, i.e., decaying to zero and sometimes bouncing back (with a positive velocity). However, the impactor velocity patterns in between different subject controllers are also quite different. For example, comparing TM with AR in Fig.~\ref{fig:head_velocities}, the impactor decelerates faster in the AR cases than that in the TM ones, especially for large impact velocities. A primary cause for these differences is the design of the locomotion controller, specifically how it orchestrates its periodic motion, manages feedback control for walking-in-place, and responds to disturbances. From the perspective of the testing operator, the intricacies of these designs remain largely undisclosed. Consequently, they might display significantly varied behaviors, resulting in different impact durations. These durations, however, cannot be directly used as clear indicators of performance. As a result, it is only fair to take the impact momentum, i.e. the momentum the testing equipment generates at the moment of impact, as the testing action $\uu_e$. It may also be different from the repeatably controlled peaking velocity as the legged robot cannot maintain absolute in-place as mentioned above, which is also part of the robot's nature in design.  

Among the three locomotion controllers tested, the most effective one (TM) was able to withstand an impact momentum of $26.376$ $\mathrm{kg}\cdot\mathrm{m/s}$. Intriguingly, this figure is roughly on par with the average effective momentum of straight punches delivered by Olympic boxers, which is reported to be 
$26.506 \mathrm{kg}\cdot\mathrm{m/s}$~\cite{walilko2005biomechanics}. Although the details are not rigorously controlled and comparable such as the impact face materials, this conceptual comparison may foster new perspectives on evaluating the performance of robotic systems in terms of ``how safe is safe enough". Specifically, it introduces the possibility of drawing measurable equivalences between the performance of legged robots and human biomechanics, contributing to a more nuanced understanding of what constitutes an adequate level of safety in legged robotic applications against external disturbances.

For the second observation mentioned above, as one can see from Fig.~\ref{fig:all_tests}, there are multiple regions where impact events sharing similar impact velocities end up with different fallover outcomes (indicated by the marker size). For example, the TL-driven robot recovers from the impact velocities of 3.214 m/s and falls over against the 3.216 m/s velocity, and that is only 0.0128 $\mathrm{kg}\cdot\mathrm{m/s}$ difference in impact momentum. A similar pattern is seen with AR at notably higher impact velocities: 3.89 m/s, 4.23 m/s, and 4.26 m/s. Out of these three impacts, the robot falls over twice, but intriguingly, the two falls aren't associated with the highest velocities. Detailed visuals related to two of the three AR impacts are available in Fig.\ref{fig:AR85_02} and Fig.\ref{fig:AR95_01}. It's worth noting that while the reason behind this phenomenon is not our primary concern, the existence of such counter-intuitive cases emphasizes the challenges inherent in creating equitable and effective safety performance testing algorithms~\cite{weng2023comparability}.

\section{Summary and Future Work}\label{sec:end}
This paper has presented a proposal of adapting the well-adopted impact generation equipment, namely the \emph{linear impactor}, as a candidate testing equipment to achieve the standardized performance testing of legged robot locomotion against external physical disturbances, i.e. the impact. We've also conducted initial laboratory tests with a linear pneumatic impactor on the Digit humanoid robot with three different locomotion controllers. These experiments have revealed interesting insights, challenges, and lessons, all contributing to an expected future of standardized safety performance testing for legged robots. 

Given this being a preliminary study, there are multiple directions of future interest including a more rigorous statistical analysis with increased tests, the expansion to other locomotion functions (e.g. speed-tracking) along with other makes and types of legged robots, the cyber-physical integration with advanced testing algorithms, and the physical enhancement of the linear impactor that better adapts to the specific needs of legged robot testing under various conditions.

\section*{Acknowledgment}
The authors thank the assistance from the students and staff at Injury Biomechanics Research Center and Cyberbotics Lab members at The Ohio State University (OSU), including Angelo Marcallini, Rosalie Connell, Zachary Haverfield, Sankalp Agrawal, etc., and the support from Michelle Murach at Transportation Research Center Inc. (TRC) and Dr. Kevin Moorhouse from National Highway Traffic Safety Administration (NHTSA). Positions and opinions advanced in this work are those of the authors and not necessarily those of OSU, TRC, or NHTSA. Responsibility for the content of the work lies solely with the authors.

\bibliographystyle{IEEEtran}
\bibliography{mybibfile}

\begin{thebibliography}{10}
\providecommand{\url}[1]{#1}
\csname url@samestyle\endcsname
\providecommand{\newblock}{\relax}
\providecommand{\bibinfo}[2]{#2}
\providecommand{\BIBentrySTDinterwordspacing}{\spaceskip=0pt\relax}
\providecommand{\BIBentryALTinterwordstretchfactor}{4}
\providecommand{\BIBentryALTinterwordspacing}{\spaceskip=\fontdimen2\font plus
\BIBentryALTinterwordstretchfactor\fontdimen3\font minus
  \fontdimen4\font\relax}
\providecommand{\BIBforeignlanguage}[2]{{%
\expandafter\ifx\csname l@#1\endcsname\relax
\typeout{** WARNING: IEEEtran.bst: No hyphenation pattern has been}%
\typeout{** loaded for the language `#1'. Using the pattern for}%
\typeout{** the default language instead.}%
\else
\language=\csname l@#1\endcsname
\fi
#2}}
\providecommand{\BIBdecl}{\relax}
\BIBdecl

\bibitem{hobbelen2007disturbance}
D.~G. Hobbelen and M.~Wisse, ``A disturbance rejection measure for limit cycle
  walkers: The gait sensitivity norm,'' \emph{IEEE Transactions on robotics},
  vol.~23, no.~6, pp. 1213--1224, 2007.

\bibitem{diedam2008online}
H.~Diedam, D.~Dimitrov, P.-B. Wieber, K.~Mombaur, and M.~Diehl, ``Online
  walking gait generation with adaptive foot positioning through linear model
  predictive control,'' in \emph{2008 IEEE/RSJ International Conference on
  Intelligent Robots and Systems}.\hskip 1em plus 0.5em minus 0.4em\relax IEEE,
  2008, pp. 1121--1126.

\bibitem{posa2017balancing}
M.~Posa, T.~Koolen, and R.~Tedrake, ``Balancing and step recovery capturability
  via sums-of-squares optimization,'' in \emph{Proceedings of Robotics: Science
  and Systems}, Cambridge, Massachusetts, July 2017.

\bibitem{meduri2023biconmp}
A.~Meduri, P.~Shah, J.~Viereck, M.~Khadiv, I.~Havoutis, and L.~Righetti,
  ``Biconmp: A nonlinear model predictive control framework for whole body
  motion planning,'' \emph{IEEE Transactions on Robotics}, vol.~39, no.~2, pp.
  905--922, 2023.

\bibitem{chen2023quadruped}
H.~Chen, Z.~Hong, S.~Yang, P.~M. Wensing, and W.~Zhang, ``Quadruped
  capturability and push recovery via a switched-systems characterization of
  dynamic balance,'' \emph{IEEE Transactions on Robotics}, vol.~39, pp.
  2111--2130, 2023.

\bibitem{castillo2021robust}
G.~A. Castillo, B.~Weng, W.~Zhang, and A.~Hereid, ``Robust feedback motion
  policy design using reinforcement learning on a 3d digit bipedal robot,'' in
  \emph{2021 IEEE/RSJ International Conference on Intelligent Robots and
  Systems (IROS)}.\hskip 1em plus 0.5em minus 0.4em\relax IEEE, 2021, pp.
  5136--5143.

\bibitem{miki2022learning}
T.~Miki, J.~Lee, J.~Hwangbo, L.~Wellhausen, V.~Koltun, and M.~Hutter,
  ``Learning robust perceptive locomotion for quadrupedal robots in the wild,''
  \emph{Science Robotics}, vol.~7, no.~62, p. eabk2822, 2022.

\bibitem{hwangbo2019learning}
J.~Hwangbo, J.~Lee, A.~Dosovitskiy, D.~Bellicoso, V.~Tsounis, V.~Koltun, and
  M.~Hutter, ``Learning agile and dynamic motor skills for legged robots,''
  \emph{Science Robotics}, vol.~4, no.~26, p. eaau5872, 2019.

\bibitem{radosavovic2023learning}
I.~Radosavovic, T.~Xiao, B.~Zhang, T.~Darrell, J.~Malik, and K.~Sreenath,
  ``Learning humanoid locomotion with transformers,'' \emph{arXiv preprint
  arXiv:2303.03381}, 2023.

\bibitem{weng2023comparability}
B.~Weng, G.~A. Castillo, W.~Zhang, and A.~Hereid, ``On the comparability and
  optimal aggressiveness of the adversarial scenario-based safety testing of
  robots,'' \emph{IEEE Transactions on Robotics}, vol.~39, pp. 3299--3318,
  2023.

\bibitem{weng2022safety}
------, ``On safety testing, validation, and characterization with
  scenario-sampling: A case study of legged robots,'' in \emph{2022 IEEE/RSJ
  International Conference on Intelligent Robots and Systems (IROS)}.\hskip 1em
  plus 0.5em minus 0.4em\relax IEEE, 2022, pp. 5179--5186.

\bibitem{aller2019state}
F.~Aller, D.~Pinto-Fernandez, D.~Torricelli, J.~L. Pons, and K.~Mombaur, ``From
  the state of the art of assessment metrics toward novel concepts for humanoid
  robot locomotion benchmarking,'' \emph{IEEE Robotics and Automation Letters},
  vol.~5, no.~2, pp. 914--920, 2019.

\bibitem{walsh2011influence}
E.~S. Walsh, P.~Rousseau, and T.~B. Hoshizaki, ``The influence of impact
  location and angle on the dynamic impact response of a hybrid iii headform,''
  \emph{Sports Engineering}, vol.~13, pp. 135--143, 2011.

\bibitem{allison2015measurement}
M.~A. Allison, Y.~S. Kang, M.~R. Maltese, J.~H. Bolte, and K.~B. Arbogast,
  ``Measurement of hybrid iii head impact kinematics using an accelerometer and
  gyroscope system in ice hockey helmets,'' \emph{Annals of biomedical
  engineering}, vol.~43, pp. 1896--1906, 2015.

\bibitem{kang2023comparison}
Y.-S. Kang, A.~Bendig, J.~Stammen, E.~Hutter, K.~Moorhouse, J.~H. Bolte~IV, and
  A.~M. Agnew, ``Comparison of small female pmhs thoracic responses to scaled
  response corridors in a frontal hub impact,'' \emph{Traffic Injury
  Prevention}, vol.~24, no.~1, pp. 62--68, 2023.

\bibitem{edwards2023kinematic}
E.~D. Edwards, T.~Landry, M.~Jesunathadas, T.~A. Plaisted, R.~J. Neice, T.~E.
  Gould, M.~Kleinberger, and S.~G. Piland, ``Kinematic assessment of the nocsae
  headform during blunt impacts with a pneumatic linear impactor,''
  \emph{Sports Engineering}, vol.~26, no.~1, p.~13, 2023.

\bibitem{meyers2008mechanical}
M.~A. Meyers and K.~K. Chawla, \emph{Mechanical behavior of materials}.\hskip
  1em plus 0.5em minus 0.4em\relax Cambridge university press, 2008.

\bibitem{varghese2003test}
J.~Varghese and A.~Dasgupta, ``Test methodology for impact testing of portable
  electronic products,'' in \emph{ASME International Mechanical Engineering
  Congress and Exposition}, vol. 37149, 2003, pp. 71--77.

\bibitem{viano2012change}
D.~C. Viano and D.~Halstead, ``Change in size and impact performance of
  football helmets from the 1970s to 2010,'' \emph{Annals of biomedical
  engineering}, vol.~40, pp. 175--184, 2012.

\bibitem{burgin2007drop}
L.~V. Burgin and R.~M. Aspden, ``A drop tower for controlled impact testing of
  biological tissues,'' \emph{Medical engineering \& physics}, vol.~29, no.~4,
  pp. 525--530, 2007.

\bibitem{cheng2010drop}
M.~Cheng, J.-P. Dionne, and A.~Makris, ``On drop-tower test methodology for
  blast mitigation seat evaluation,'' \emph{International Journal of Impact
  Engineering}, vol.~37, no.~12, pp. 1180--1187, 2010.

\bibitem{li2018development}
W.~Li, S.~Li, L.~Yan, D.~Qin, S.~Sun, and H.~Zhao, ``Development and field
  application of a pulse-jet hydraulic impactor,'' \emph{Natural Gas Industry
  B}, vol.~5, no.~6, pp. 558--564, 2018.

\bibitem{mejia2019linear}
C.~H. Mejia, J.~Jayet, P.~Germano, A.~Thabuis, and Y.~Perriard, ``Linear impact
  generator for automated dataset acquisition of elastic waves in haptic
  surfaces,'' in \emph{2019 22nd International Conference on Electrical
  Machines and Systems (ICEMS)}.\hskip 1em plus 0.5em minus 0.4em\relax IEEE,
  2019, pp. 1--5.

\bibitem{rao2019test}
S.~J. Rao, G.~J. Forkenbrock \emph{et~al.}, ``Test procedures traffic jam
  assist test development considerations,'' United States. Department of
  Transportation. National Highway Traffic Safety Administration, Tech. Rep.,
  2019.

\bibitem{castillo2023template}
G.~A. Castillo, B.~Weng, S.~Yang, W.~Zhang, and A.~Hereid, ``Template model
  inspired task space learning for robust bipedal locomotion,'' in \emph{2023
  IEEE/RSJ International Conference on Intelligent Robots and Systems
  (IROS)}.\hskip 1em plus 0.5em minus 0.4em\relax IEEE, 2023.

\bibitem{del2016implementing}
A.~Del~Prete, N.~Mansard, O.~E. Ramos, O.~Stasse, and F.~Nori, ``Implementing
  torque control with high-ratio gear boxes and without joint-torque sensors,''
  \emph{International Journal of Humanoid Robotics}, vol.~13, no.~01, p.
  1550044, 2016.

\bibitem{raibert1983dynamically}
\BIBentryALTinterwordspacing
M.~H. Raibert, J.~Brown, C.~H. B., K.~Michael, and H.~Jeff, ``Dynamically
  stable legged locomotion,'' 1989. [Online]. Available:
  \url{https://dspace.mit.edu/handle/1721.1/6820}
\BIBentrySTDinterwordspacing

\bibitem{lee2022online}
J.~Lee, J.~Ahn, D.~Kim, S.~H. Bang, and L.~Sentis, ``Online gain adaptation of
  whole-body control for legged robots with unknown disturbances,''
  \emph{Frontiers in Robotics and AI}, vol.~8, p. 788902, 2022.

\bibitem{romualdi2022online}
G.~Romualdi, S.~Dafarra, G.~L'Erario, I.~Sorrentino, S.~Traversaro, and
  D.~Pucci, ``Online non-linear centroidal mpc for humanoid robot locomotion
  with step adjustment,'' in \emph{2022 International Conference on Robotics
  and Automation (ICRA)}.\hskip 1em plus 0.5em minus 0.4em\relax IEEE, 2022,
  pp. 10\,412--10\,419.

\bibitem{walilko2005biomechanics}
T.~J. Walilko, D.~C. Viano, and C.~A. Bir, ``Biomechanics of the head for
  olympic boxer punches to the face,'' \emph{British journal of sports
  medicine}, vol.~39, no.~10, pp. 710--719, 2005.

\end{thebibliography}

\end{document}